\documentclass[runningheads]{llncs}
\usepackage{graphicx}

\usepackage{tikz}
\usepackage{comment}
\usepackage{amsmath,amssymb} 
\usepackage{color}
\usepackage{bm}
\usepackage{arydshln}
\usepackage{float}
\usepackage{setspace}
\usepackage{subfigure}

\usepackage[accsupp]{axessibility}  
\usepackage{bbding}

\usepackage{multirow}

\begin{document}
\pagestyle{headings}
\mainmatter
\def\ECCVSubNumber{1441}  

\title{Static and Dynamic Concepts for Self-supervised Video Representation Learning} 

\titlerunning{Video Concept Learning}
%
\author{Rui Qian\inst{1} \and
Shuangrui Ding\inst{2} \and
Xian Liu\inst{1}
\and
Dahua Lin\inst{1,3}\thanks{Corresponding author. Email: dhlin@ie.cuhk.edu.hk}}
\authorrunning{R. Qian et al.}
%
\institute{The Chinese University of Hong Kong, Hong Kong, China \and
Shanghai Jiao Tong University, Shanghai, China \and
Shanghai Artificial Intelligence Laboratory, Shanghai, China
\email{\{qr021,lx021,dhlin\}@ie.cuhk.edu.hk dsr1212@sjtu.edu.cn}}
\maketitle

\begin{abstract}
In this paper, we propose a novel learning scheme for self-supervised video representation learning. Motivated by how humans understand videos, we propose to first learn general visual concepts then attend to discriminative local areas for video understanding. Specifically, we utilize static frame and frame difference to help decouple static and dynamic concepts, and respectively align the concept distributions in latent space. We add diversity and fidelity regularizations to guarantee that we learn a compact set of meaningful concepts. Then we employ a cross-attention mechanism to aggregate detailed local features of different concepts, and filter out redundant concepts with low activations to perform local concept contrast. Extensive experiments demonstrate that our method distills meaningful static and dynamic concepts to guide video understanding, and obtains state-of-the-art results on UCF-101, HMDB-51, and Diving-48.
\end{abstract}

\section{Introduction}

\begin{figure}[t]
    \centering
    \subfigure[Soccer Juggling]{
    \includegraphics[width=0.15\linewidth]{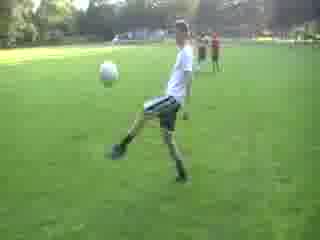}
    \includegraphics[width=0.15\linewidth]{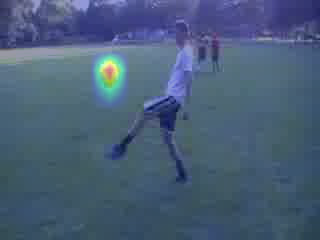}
    \includegraphics[width=0.15\linewidth]{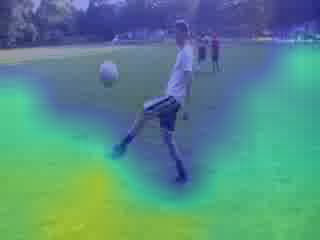}
    \includegraphics[width=0.15\linewidth]{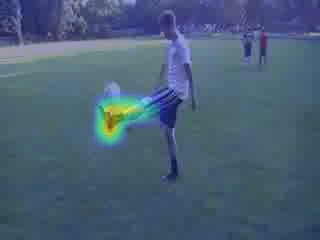}
    \includegraphics[width=0.15\linewidth]{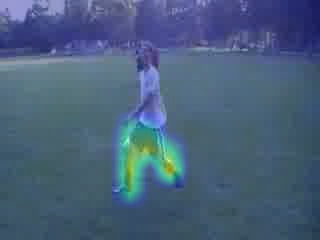}
    \includegraphics[width=0.15\linewidth]{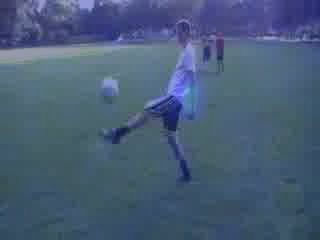}}\\
    \subfigure[Basketball]{
    \includegraphics[width=0.15\linewidth]{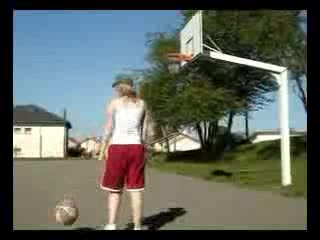}
    \includegraphics[width=0.15\linewidth]{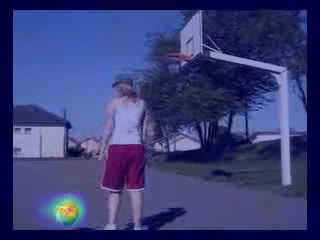}
    \includegraphics[width=0.15\linewidth]{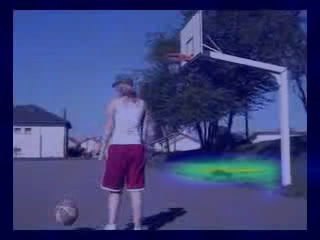}
    \includegraphics[width=0.15\linewidth]{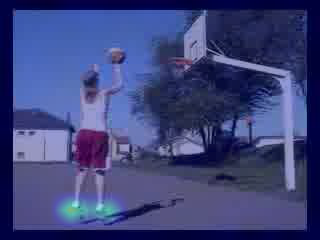}
    \includegraphics[width=0.15\linewidth]{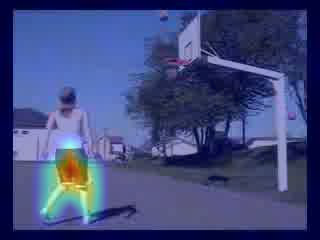}
    \includegraphics[width=0.15\linewidth]{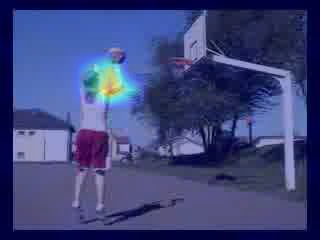}}
    \caption{Visualization of visual concept attention maps. Each column corresponds to the same concept, the former two columns describe static concepts and the latter three present dynamic concepts. The same visual concept highlights similar visual patterns, e.g., spherical objects, grass land, foot movement, leg movement, arm movement.}
    \label{attention}
\end{figure}

Self-supervised representation learning has been an exciting problem in computer vision, which aims to encode robust representations that can be transferred to various downstream tasks without human labeling. A prevalent strategy is to design pretext tasks and acquire pseudo labels as self-supervision~\cite{caron2018deep,gidaris2018unsupervised} or employ contrastive learning to discriminate instances~\cite{chen2020simple,he2020momentum,caron2020unsupervised}. However, this learning scheme is inconsistent with how humans learn from the world. To be specific, instead of solely learning from labels or contrasting global features, humans can typically conclude some general basic concepts from detailed observations, then make predictions based on these concepts~\cite{bucher2018semantic,seel2011encyclopedia,koh2020concept}. For example, we identify an airplane through its wings and rudder; and recognize the action of playing soccer through the ball as well as running and kicking movement as in Fig~\ref{attention}. To this end, it would be promising to automatically formulate transferable concepts to guide detailed local feature perception and improve the representations.

There have been some works exploring learning interpretable visual concepts for particular tasks~\cite{koh2020concept,bucher2018semantic,xiong2021explore,chen2020concept}. But in unsupervised video representation learning, how to formulate meaningful visual concepts and efficiently leverage local cues remains unsolved. The difficulty lies in two aspects: Videos contain more redundancy on temporal dimension. Besides, we lack fine-grained supervision on the potential visual concepts. Most of the recent state-of-the-art works on video representation learning inherit contrastive learning framework~\cite{qian2020spatiotemporal,han2020self,feichtenhofer2021large}, which projects the global pooled feature vectors into a latent space and performs instance discrimination. Compared with the aforementioned human perception, this formulation explicitly contrasts high-level global feature vectors but has difficulty dealing with detailed local features. Some works propose region-based local feature contrast but could result in high redundancy~\cite{wang2021dense,yuan2021contextualized}. In order to effectively utilize the detailed local features, we propose a novel learning strategy for self-supervised video representation learning. We aggregate local features that present similar concepts, and then perform the concept-level alignment.

Concretely, we propose to form a latent space consisting of the learned visual concepts, and leverage the latent concept distributions as self-supervision to jointly optimize feature representations and concept descriptions. However, since the feature attributes are highly entangled in the high-level representation, it is nontrivial to directly obtain general concepts without annotations. To solve this, we divide the learning concepts into two general divisions, i.e., static scenes and dynamic motions. Those two concepts are proved to be complementary but orthogonal for video representation learning~\cite{huang2021self}. Static scenes focus on background cues while dynamic motions lay more emphasis on object's movement. In practice, we use the simple static frame and frame difference to naturally decouple these two aspects and ameliorate the entanglement of high-level feature. 
Further, we define the projection head as a cosine classifier to generate concept latent codes, with each class corresponding to a potential static or dynamic local concept. 
We respectively align the static (dynamic) concept latent codes between original video and static frame (frame difference), and encourage sparsity in the latent space to guarantee diversity of learned concepts. 
Besides, to make the projection head preserve necessarily relevant information and reduce redundancy, we regard the latent codes as information bottleneck, where they are expected to reconstruct the initial feature vectors. Thus, we apply a light-weight MLP to achieve the fidelity regularization. By doing so, we establish a concept-based latent space consisting of general static and dynamic visual concepts.

With these learned concept prototypes, we attend to local concepts in each spatio-temporal area to improve the detailed local feature modeling. Specifically, we use cross-attention to aggregate local features, then output a set of features belonging to different concepts like Fig.~\ref{attention}. By referring to the concept latent code, we select a series of visual concepts with high activations as valid ones and filter out the redundant feature pairs. Contrastive loss is applied to these valid pairs for fine-grained alignment. In this way, we seamlessly integrate general concept learning with detailed local feature perception to enhance video representations.

To sum up, our contributions are: (1) We propose a novel self-supervised video representation learning scheme, where we formulate general concepts to guide concept-level detailed local feature alignment. (2) We employ cross-attention to aggregate detailed features of different concepts, and filter out redundant local features by concept latent codes. In this way, we achieve efficient local concept contrast. (3) We achieve state-of-the-art results on downstream action recognition and video retrieval across UCF-101, HMDB-51 and Diving-48 datasets.

\section{Related Work}

\noindent\textbf{Self-supervised Learning.} Self-supervised learning aims to make full use of large-scale unlabelled data without resorting to human annotations. Some works design pretext tasks, e.g., image rotation~\cite{gidaris2018unsupervised}, colorization~\cite{kim2018learning}, clustering~\cite{caron2018deep,regatti2020consensus}, to obtain pseudo labels and guide representation learning. Another line of works introduce contrastive learning to build robust feature representations~\cite{wu2018unsupervised,dosovitskiy2014discriminative,oord2018representation}. They employ noise contrastive estimation~\cite{gutmann2010noise} to compare feature representations and discriminate different instances~\cite{tian2020contrastive,chen2020simple,he2020momentum}. Technically, these methods rely on nonlinear projection head to project the extracted features into a latent space for contrastive loss computation to reduce information loss. However, without explicit constraint on the projection head, what information is preserved and contrasted in the latent space is unclear, and the learning process is of low interpretability. More recently, \cite{ermolov2021whitening} employs whitening to analyze the latent feature space. \cite{caron2020unsupervised} assigns features to prototype vectors and contrasts cluster assignments in the latent space. In contrast, in this work, we enforce the projection heads to learn potential visual concepts and formulate an interpretable latent space, where we contrast the concept distributions to guide general representation learning.

\noindent\textbf{Video Representation Learning.} Representation learning in video domain requires the model to capture crucial spatio-temporal relationships in video sequences. Early works employ the temporal transformation~\cite{misra2016shuffle,xu2019self,yao2020seco,benaim2020speednet,jenni2020video,chen2021rspnet,yang2020video}, spatio-temporal jigsaw~\cite{kim2019self,wang2019self}, temporal cycle-consistency~\cite{jabri2020space,li2019joint,wang2019learning}, future prediction~\cite{behrmann2021unsupervised,luo2017unsupervised} as pretext tasks. Later, \cite{qian2020spatiotemporal,wang2020self,feichtenhofer2021large,qian2020multiple,liu2022semantic,liu2022visual,hu2020discriminative,liu2022learning} expand contrastive learning framework to video and audio-video domain. Further, \cite{jenni2021time,kuang2021video,huang2021ascnet,ding2022motion,ding2022dual} utilize the internal temporal structure to generate richer positive samples. \cite{han2020self,li2021motion,asano2020labelling,miech2020end,piergiovanni2020evolving} contrast temporally aligned multi-modal inputs to learn complementary information. These works explicitly contrast the global representations of video clips, but pay little attention to detailed local features. To this end, \cite{han2019video,han2020memory} propose to predict dense feature maps in future timestamps. \cite{recasens2021broaden,behrmann2021long,dave2021tclr,qian2021exploring} contrast short and long clips on each timestamp to attend to fine-grained temporal features, but still fail to utilize detailed spatial cues. \cite{yuan2021contextualized,chen2021previts} rely on bounding boxes or segmentation masks to align semantically related local areas. While in our work, we use simple static frame and frame difference to distill static and dynamic visual concepts, based on which we aggregate relevant information from each spatio-temporal area to enhance detailed content modeling.

\noindent\textbf{Concept Learning.} Recently, there have emerged a line of works that learn human-specified visual concepts to solve downstream visual tasks~\cite{koh2020concept,chen2020concept,bucher2018semantic,losch2019interpretability,de2018clinically}. They design concept bottlenecks models to first predict concepts then use these concepts to make final predictions. Comparing to end-to-end deep models, concept bottleneck models are more interpretable but require extra concept annotations. To tackle this problem, \cite{sawada2022concept,alvarez2018towards} develop various regularizations to constrain the concept bottleneck and obtain potential concepts. \cite{xiong2021explore} points out that one-hot category labels are not optimal concept descriptions, and devises an exploration-experience loss to alternatively update feature representation and concept description. To our best knowledge, we are the first to integrate concept learning into self-supervised video representation learning. We utilize static and dynamic visual concepts to learn both general and detailed video representations.

\section{Method}

\begin{figure}[t]
    \centering
    \includegraphics[width=0.9\linewidth]{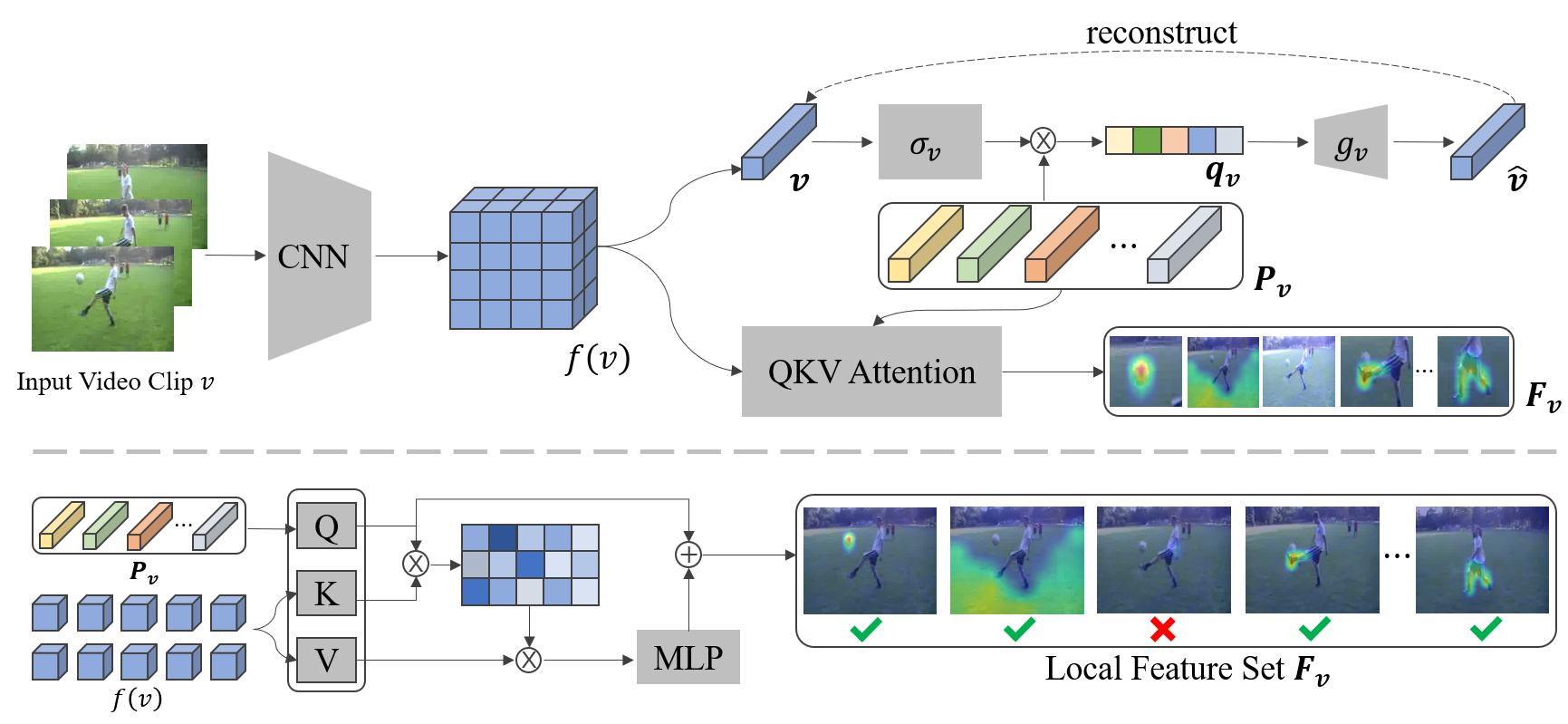}
    \caption{Overview of the framework. We take the original input video clip $v$ for illustration. In the upper branch, we calculate cosine similarity between concept prototypes $\bm{P_v}$ and transformed video feature $\sigma_v(\bm{v})$ as concept latent code $\bm{q_v}$, which is then passed through simple MLP to reconstruct the original feature vector. In the lower branch, we use QKV attention with residue to aggregate local features of different concepts and refer to $\bm{q_v}$ to avoid redundant local concept contrast.}
    \label{framework}
\end{figure}

Our framework is shown in Fig.~\ref{framework}. For simplicity, we show detailed procedures for video clip $v$, while static frame $s$ and $d$ are processed similarly. Specifically, we first propose decoupled concept alignment (Sec.~\ref{3.1}) with regularizations (Sec.~\ref{3.2}) to jointly optimize the extracted features and concept descriptions. Then referring to learned concepts, we employ cross-attention to aggregate detailed local features of different concepts, filter out redundant concepts with low activations and perform concept-level alignment (Sec.~\ref{3.3}).

\subsection{Decoupled Concept Learning}
\label{3.1}
Videos typically possess two complementary concepts, static concepts that indicate background scene attributes, and dynamic concepts that reveal human or object movements. Given a video sequence $v$, since various visual concepts are highly entangled, it is nontrivial to directly learn meaningful visual concepts without resorting to human annotations. But it is practical to decouple the static and dynamic information in the input stage, i.e., we randomly select a static frame $s$ and calculate frame difference $d$ to respectively carry static and dynamic attributes. Then, an intuitive idea is to learn potential static concepts from $v$ and $s$, extract dynamic concepts from $v$ and $d$, and respectively perform static and dynamic concept alignment.

\noindent\textbf{Concept Prototypes.} To formulate the latent concept space, we propose to learn several prototypes, each corresponding to a static or dynamic concept. Specifically, we define three sets of prototypes respectively for $s, d, v$ as:
\begin{align}
    \bm{P_s}\in\mathbb{R}^{K_s\times C}, \quad\bm{P_d}\in\mathbb{R}^{K_d\times C}, \quad\bm{P_v}\in\mathbb{R}^{(K_s+K_d)\times C},
\end{align}
where $C$ denotes channel dimension, $K_s$ is the number of static concepts, $K_d$ is the number of dynamic concepts. We use these concept prototypes to generate latent concept activation codes and retrieve relevant local features in later stage.

\noindent\textbf{Concept Codes.} Following~\cite{chen2020simple,caron2020unsupervised}, we use a projection head to project the features into a latent space and generate the concept latent codes. Mathematically, we denote the feature extractor as $f$, and employ global average pooling to obtain three feature vectors\footnote{For simplicity, we use the same symbol to denote the backbone for $s, d, v$.}:
\begin{align}
    \bm{s} = GAP(f(s)), \quad\bm{d} = GAP(f(d)), \quad\bm{v} = GAP(f(v)),
\end{align}
each is of the same dimension $\mathbb{R}^C$. Then, we pass these feature vectors through projection heads to calculate concept codes. For illustration, we take the concept code $\bm{q_s}$ for static frame as an example. We first input $\bm{s}$ into a transformation $\sigma_s$, which is in default identity mapping but can be replaced with other shallow layers like MLP. Then we calculate the cosine similarity between the output vector and each prototype to form $\bm{q_s}$:
\begin{align}
    \bm{q_s}^{(k)} = \frac{\bm{P_s}^{(k)}\sigma_s(\bm{s})^T}{||\bm{P_s}^{(k)}||_2||\sigma_s(\bm{s})||_2}, \quad\bm{q_s}\in\mathbb{R}^{K_s},
\end{align}
where the superscript $(k)$ indicates $k$-th channel. Similarly, we obtain concept codes $\bm{q_d}\in\mathbb{R}^{K_d}$ and $\bm{q_v}\in\mathbb{R}^{K_s+K_d}$ in the same manner.

\noindent\textbf{Concept Alignment.} Since the video and static frame share the same static attributes, while video and frame difference have the same dynamic attributes, we propose to respectively align the static and dynamic concepts through Eq.~\ref{concept-align}. 
\begin{align}
\begin{split}
    \mathcal{L}_{aln} = &-\sum_{k=1}^{K_s}\left(\overline{\bm{q_s}}^{(k)}\log\frac{\exp(\bm{q_v}^{\bm{s}(k)}/\tau)}{\sum_{k'}\exp(\bm{q_v}^{\bm{s}(k')}/\tau)}+\overline{\bm{q_v^s}}^{(k)}\log\frac{\exp(\bm{q_s}^{(k)}/\tau)}{\sum_{k'}\exp(\bm{q_s}^{(k')}/\tau)}\right)\\
    &-\sum_{k=1}^{K_d}\left(\overline{\bm{q_d}}^{(k)}\log\frac{\exp(\bm{q_v}^{\bm{d}(k)}/\tau)}{\sum_{k'}\exp(\bm{q_v}^{\bm{d}(k')}/\tau)}+\overline{\bm{q_v^d}}^{(k)}\log\frac{\exp(\bm{q_d}^{(k)}/\tau)}{\sum_{k'}\exp(\bm{q_d}^{(k')}/\tau)}\right),\\
    \label{concept-align}
\end{split}
\end{align}
For simplicity, we divide $\bm{q_v}$ into two parts, the former $K_s$ channels as $\bm{q_v^s}$ indicating static concepts, and the latter $K_d$ channels as $\bm{q_v^d}$ for dynamic concepts. Similar to SWAV~\cite{caron2020unsupervised}, we assume the concepts follow a uniform distribution over the whole dataset, and use Sinkhorn-Knopp algorithm~\cite{cuturi2013sinkhorn} to generate the soft code $\overline{\bm{q}}$. Then we calculate the cross-entropy between $\overline{\bm{q}}$ and the latent concept distribution by taking softmax with temperature $\tau$ on $\bm{q}$. By minimizing $\mathcal{L}_{aln}$, we respectively align static and dynamic concept distributions, and jointly optimize feature representations and concept descriptions from large-scale video data.

\subsection{Concept Bottleneck Constraint}
\label{3.2}
However, the decoupled concept alignment objective alone cannot guarantee that each of the learned prototype corresponds to a meaningful concept. Motivated by~\cite{alvarez2018towards}, the general concepts should possess fidelity and diversity. That is, the concepts should preserve much relevant information from the inputs, and the inputs can be described by a few concepts. To this end, we devise two constraints on the concept latent codes as follows.

The first constraint is the sparsity regularization term as Eq.~\ref{diversity} to enforce diversity of learned concepts. We employ $\mathcal{L}_1$ norm regularization to encourage sparsity of concept latent codes, so that each input activates only a few concepts.
\begin{align}
    \mathcal{L}_{div} = \left\lVert\bm{q_s}\right\rVert_1+\left\lVert\bm{q_d}\right\rVert_1+\left\lVert\bm{q_v}\right\rVert_1.
    \label{diversity}
\end{align}
The second constraint is a reconstruction loss as Eq.~\ref{fidelity} to ensure fidelity and reduce redundancy. We borrow the idea from autoencoder to reconstruct the feature vectors. Since the channel dimension of concept code is smaller than the feature vector, we regard $\bm{q}$ as information bottleneck and pass them through two-layer MLP $g$ for reconstruction. We use $\mathcal{L}_2$ loss for optimization, and stop gradient on the original features. In this way, the concept prototypes cover a wide range of important information with low redundancy.
\begin{align}
    \mathcal{L}_{fid} = \left\lVert g_s(\bm{q_s})-\bm{s}\right\rVert_2^2+\left\lVert g_d(\bm{q_d})-\bm{d}\right\rVert_2^2+\left\lVert g_v(\bm{q_v})-\bm{v}\right\rVert_2^2.
    \label{fidelity}
\end{align}

\noindent\textbf{Relation to SWAV.} Our concept code formulation is similar to SWAV~\cite{caron2020unsupervised}, both using cosine similarity between feature vectors and prototypes. But the motivations and technical designs are different. In terms of the motivation, SWAV is essentially over-clustering and the prototypes are cluster centroids, the number of which is set as 3,000 in default, much greater than semantic categories. While in our method, the prototypes project the feature vectors into the low dimensional space, which interprets the concept activations instead of the instance discrimination. Through regularizations and activation alignment, our prototypes are an ordered set of interpretable concepts each presenting a visual attribute. In terms of technical design, our method only conducts spatio-temporal cropping due to multiple modalities while SWAV requires stronger augmentation to make the pretraining task harder and improve the representation quality.

\subsection{Local Concept Contrast}
\label{3.3}
The global concept code alignment serves as an effective supervision to learn spatio-temporal characters in videos, but does not make use of detailed local features which are crucial for video understanding. Some existing works in image domain first match corresponding local areas then make contrast~\cite{wang2021dense,xie2021propagate}, but they have difficulty expanding to videos because of the redundancy on time dimension. \cite{yuan2021contextualized} employs bounding boxes for region-based contrast between video clips, but requires prior to filter redundant background areas. In order to better utilize the detailed local contents, we need to generate a compact set of local features with low redundancy. Therefore, we propose to leverage the learned prototypes to retrieve detailed local features that are relevant to particular concepts, and output an ordered set of local features for effective contrast.

\noindent\textbf{Local Feature Attention.} Motivated by the success of attention mechanism in local feature aggregation~\cite{vaswani2017attention,dosovitskiy2020image,wang2018non,gao2021container}, we employ widely used cross-attention mechanism to retrieve detailed local features that are relevant to specific visual concepts. As illustrated in Fig.~\ref{framework}, we linearly project the concept prototypes as query tokens, and project the feature maps to formulate key and value tokens. Then QKV attention with residue is applied to aggregate local features related to the query. We still use the local features on static frame as an example:
\begin{align}
    \bm{F_s} = QKV(\bm{P_s}, f(\bm{s}), f(\bm{s})), \quad\bm{F_s}\in\mathbb{R}^{K_s\times C}.
\end{align}
We obtain $\bm{F_d}\in\mathbb{R}^{K_d\times C}$ and $\bm{F_v}\in\mathbb{R}^{(K_s+K_d)\times C}$ in the same manner. Similar to the separation on $\bm{q_v}$, we also divide $\bm{F_v}$ into $\bm{F_v^s}\in\mathbb{R}^{K_s\times C}$ and $\bm{F_v^d}\in\mathbb{R}^{K_d\times C}$.

Since each prototype corresponds to a potential static or dynamic concept, each generated attention map highlights local areas that contain particular concepts as shown in Fig.~\ref{attention}, where each column belongs to the same concept. Therefore, it is intuitive to apply contrastive loss on the aggregated features of the matching concepts to further enhance detailed local representations.

\noindent\textbf{Local Feature Contrast.} Recall that each input is representable with a few concepts, we need to first filter out a set of valid concepts that exist in the input sample. To do this, we resort to the previously obtained concept latent codes $\bm{q}$, which figure out which concepts are activated in each training sample. Mathematically, we take local features of static concepts for illustration. Given concept latent codes $\bm{q_s}, \bm{q_v^s}$ and local features $\bm{F_s}, \bm{F_v^s}$, we select top-$K$ indexes of each latent code and take the intersection as the valid static concept indexes:
\begin{align}
    \bm{idx_s} = \text{top-k}(\bm{q_s}, K)\cap\text{top-k}(\bm{q_v^s}, K).
\end{align}
The valid local feature pairs are denoted as $\{(\bm{F_s}^{(k)}, \bm{F_v}^{s(k)}) | k\in\bm{idx_s}\}$, with the superscript $(k)$ indicating local feature of $k$-th concept.

These local features of the same static (dynamic) concept from the same video are expected to represent exactly the same appearances (movements), thus should be aligned. To this end, we apply contrastive margin loss in Eq.~\ref{local-static} to contrast the local features of valid concept indexes. To be specific, we employ the valid local feature pair from the same video as positive samples, and use local features of corresponding concept from other videos in the mini-batch to form negative samples. We minimize the $\mathcal{L}_2$ distance between positive feature pairs, and push the distance between negative pairs to a large margin:
\begin{align}
    l(\bm{F_s}, \bm{F_v^s}) = \sum_{k\in\bm{idx_s}}\left[\left\lVert\bm{F_s}^{(k)}-\bm{F_v}^{s(k)}\right\lVert_2^2+\sum_{\widetilde{\bm{F}}\in\mathcal{N}}\max\left(\lambda-\left\lVert\bm{F_s}^{(k)}-\bm{\widetilde{F}_v}^{s(k)}\right\lVert_2, 0\right)^2\right],
    \label{local-static}
\end{align}
where $\lambda$ is the margin hyper-parameter, and $\mathcal{N}$ is the set of negative samples in the mini-batch. We use similar techniques to process local features of dynamic concepts, and the final local concept contrast learning objective is formulated as
\begin{align}
    \mathcal{L}_{loc} = l(\bm{F_s}, \bm{F_v^s})+l(\bm{F_v^s}, \bm{F_s})+l(\bm{F_d}, \bm{F_v^d})+l(\bm{F_v^d}, \bm{F_d}).
\end{align}
By minimizing $\mathcal{L}_{loc}$, we build a concept-level self-supervision to make use of detailed local features and improve video representations. Comparing to previous methods using similar techniques to contrast local features~\cite{yuan2021contextualized,weinzaepfel2022learning}, our method does not rely on prior or complex post-processing to filter out redundant feature pairs. The integration of general concept learning and detailed local feature contrast leads to higher learning efficiency and more comprehensive representations.

\noindent\textbf{Overall Learning Objective.} The overall training objective can be written as
\begin{align}
    \mathcal{L} = \mathcal{L}_{aln} + \alpha\mathcal{L}_{loc} + \beta\mathcal{L}_{fid} + \gamma\mathcal{L}_{div},
\end{align}
where the balancing hyper-parameters are respectively set to $\alpha=\beta=1, \gamma=0.01$ in default. Since the formulation of $\mathcal{L}_{loc}$ relies on the concept codes to filter out valid pairs, in the first few epochs (5 epochs in default), we do not include $\mathcal{L}_{loc}$ to prevent random selection and stabilize training.

\section{Experiment}

\subsection{Dataset}

We use 4 popular video datasets, Kinetics-400~\cite{carreira2017quo}, UCF-101~\cite{soomro2012ucf101},  HMDB-51~\cite{kuehne2011hmdb} and Diving-48~\cite{li2018resound}. \textbf{Kinetics-400}~\cite{carreira2017quo} is a widely used benchmark for self-supervised video representation learning, with 240K video clips covering 400 human action classes. \textbf{UCF-101}~\cite{soomro2012ucf101} covers 101 action categories and more than 13K annotated clips. \textbf{HMDB-51}~\cite{kuehne2011hmdb} contains around 7k clips covering 51 action classes. \textbf{Diving-48}~\cite{li2018resound} contains 48 different diving actions. Different action classes in Diving-48 mainly vary in motion patterns and the backgrounds are quite similar. 

\subsection{Implementation Details}

We choose R(2+1)D-18~\cite{tran2018closer} with 14.4M parameters, and S3D~\cite{xie2018rethinking} as the video encoder. We empirically find that using separate networks or sharing the same network to extract RGB/static frame/frame difference features leads to similar performance. But using shared backbone results in higher learning efficiency, so we use the same backbone for all in default. Given a video clip, we randomly select a frame and repeat 16 times on the temporal axis to construct static frame input, and use the difference between adjacent frames to form the frame difference input. The resolution of each input sequence is $16\times 112\times 112$ if not specially motioned. We pretrain the model for 200 epochs on UCF-101 or 100 epochs on Kinetics-400. We adopt SGD optimizer with the initial learning rate of $10^{-2}$ and weight decay of $10^{-4}$. We set the number of static or dynamic concepts to $K_s=K_d=50$, and the ratio of valid local concepts to 10\%, $K=5$ in default.



\begin{table}[t]
\centering
\small
\scalebox{0.9}{
    \begin{tabular}{c|ccccc|cc}
        \hline
        Method & Backbone & Pretrain Dataset & Frames & Res. & Freeze & UCF-101 & HMDB-51 \\\hline
        CBT~\cite{sun2019learning} & S3D & Kinetics-600 & 16 & 112 & \Checkmark & 54.0 & 29.5 \\
        RSPNet~\cite{chen2021rspnet} & R3D & Kinetics-400 & 16 & 112 & \Checkmark & 61.8 & 42.8 \\
        MLRep~\cite{qian2021enhancing} & R3D & Kinetics-400 & 16 & 112 & \Checkmark & 63.2 & 33.4 \\
        CoCLR$\dagger$~\cite{han2020self} & S3D & Kinetics-400 & 32 & 128 & \Checkmark & 74.5 & 46.1 \\
        \hdashline
        \textbf{Ours} & R(2+1)D & UCF-101 & 16 & 112 & \Checkmark & 67.4 & 40.7 \\
        \textbf{Ours} & R(2+1)D & Kinetics-400 & 16 & 112 & \Checkmark & 72.1 & 45.9 \\
        \textbf{Ours} & S3D & Kinetics-400 & 16 & 128 & \Checkmark & 75.1 & 47.4 \\
        \hline
        \hline
       
        TempTrans~\cite{jenni2020video} & R(2+1)D & UCF-101 & 16 & 112 & \XSolidBrush & 81.6 & 46.4 \\
        LSFD~\cite{behrmann2021long} & R3D & UCF-101 & 32 & 112 & \XSolidBrush & 77.2 & 53.7 \\
        STS$\dagger$~\cite{wang2020statistic} & R(2+1)D & UCF-101 & 16 & 112 & \XSolidBrush & 77.8 & 40.7 \\
        CoCLR$\dagger$~\cite{han2020self} & S3D & UCF-101 & 32 & 128 & \XSolidBrush & 81.4 & 52.1 \\
        \hdashline
        \textbf{Ours} & R(2+1)D & UCF-101 & 16 & 112 & \XSolidBrush & 82.1 & 49.7 \\
        \textbf{Ours} & S3D & UCF-101 & 32 & 128 & \XSolidBrush & 83.7 & 53.8 \\
        \hline
        ASCNet~\cite{huang2021ascnet} & R3D & Kinetics-400 & 16 & 112 & \XSolidBrush & 80.5 & 52.3 \\
        Pace~\cite{wang2020self} & R(2+1)D & Kinetics-400 & 16 & 112 & \XSolidBrush & 77.1 & 36.6
        \\
        VideoMoCo~\cite{pan2021videomoco} & R(2+1)D & Kinetics-400 & 32 & 112 & \XSolidBrush & 78.7 & 49.2
        \\
        RSPNet~\cite{chen2021rspnet} & R(2+1)D & Kinetics-400 & 16 & 112 & \XSolidBrush & 81.1 &44.6 \\
        TCLR~\cite{dave2021tclr} & R(2+1)D & Kinetics-400 & 16 & 112 & \XSolidBrush & 84.3 & 54.2 \\
        TimeEq~\cite{jenni2021time} & S3D-G & Kinetics-400 & 32 & 128 & \XSolidBrush & 86.9 & 63.5 \\
        STS$\dagger$~\cite{wang2020statistic} & S3D-G & Kinetics-400 & 64 & 224 & \XSolidBrush & 89.0 & 62.0 \\
        CoCLR$\dagger$~\cite{han2020self} & S3D & Kinetics-400 & 32 & 128 & \XSolidBrush & 87.9 & 54.6 \\
        \hdashline
        \textbf{Ours} & R(2+1)D & Kinetics-400 & 16 & 112 & \XSolidBrush & 86.1 & 54.8 \\
        \textbf{Ours} & S3D & Kinetics-400 & 16 & 128 & \XSolidBrush & 88.3 & 56.4 \\
        \hline
    \end{tabular} 
    }
    \caption{Results on action recognition downstream task. We present the backbone encoder, pretrain dataset, spatio-temporal resolution of each method. Freeze (tick) indicates \textit{linear probe}, and no freeze (cross) denotes \textit{end-to-end finetune}.}
    \label{tab:recognition}
\end{table}

\subsection{Evaluation on Downstream Tasks}

\noindent\textbf{Action Recognition.}
We first present action recognition in Table~\ref{tab:recognition}. We report \textit{linear probe} and \textit{finetune} Top-1 accuracy. For fair comparison, we exclude the works with different evaluation settings and much deeper backbone~\cite{qian2020spatiotemporal,feichtenhofer2021large} or rely on audio and text~\cite{recasens2021broaden,miech2020end}. The $\dagger$ means jointly utilizing RGB and optical flow for pretraining, and the final performance is tested with RGB only.

In \textit{linear probe} settings, our method achieves state-of-the art results on both two datasets. It is worth noting that our UCF-101 pretrained model even outperforms most RGB-based methods pretrained on Kinetics-400, which indicates the high data efficiency of our learning framework. Regarding to comparison with CoCLR~\cite{han2020self} pretrained with RGB and Flow, we reach higher accuracy with fewer frames in each clip. It indicates that simple frame difference could replace computationally expensive optical flow to improve dynamic attribute learning. 

In \textit{finetune}, ours also achieves the best results among RGB-only methods, and is comparable with RGB-Flow two-stream models. Among these method, \cite{jenni2020video,jenni2021time,chen2021rspnet,huang2021ascnet} carefully design temporal transformations to enhance temporal perception in videos, \cite{behrmann2021long,dave2021tclr} employ short and long clips to attend to fine-grained temporal features, \cite{han2020self,wang2020statistic} utilize complementary information between RGB and Flow to enhance video representations. While our method proposes to formulate general static and dynamic concepts to guide detailed local feature perception, the performance demonstrates the effectiveness of our new learning scheme.

\begin{table}[t]
    \centering
    \small
    \begin{tabular}{c|c|cccc|cccc}
        \hline
        \multirow{2}{*}{Method} & \multirow{2}{*}{Backbone} & \multicolumn{4}{c|}{UCF-101} & \multicolumn{4}{c}{HMDB-51} \\
        \cline{3-10}
         & & R@1 & R@5 & R@10 & R@20 & R@1 & R@5 & R@10 & R@20 \\
        \hline
        VCP~\cite{luo2020video} & R3D & 18.6 & 33.6 & 42.5 & 53.3 & 7.6 & 24.4 & 36.3 & 53.6 \\
        MLRep~\cite{qian2021enhancing} & R3D & 39.6 & 57.6 & 69.2 & 78.0 & 18.8 & 39.2 & 51.0 & 63.7 \\
        VCLR~\cite{kuang2021video} & R2D-50 & 46.8 & 61.8 & 70.4 & 79.0 & 17.6 & 38.6 & 51.1 & 67.6 \\
        PRP~\cite{yao2020video} & R(2+1)D & 20.3 & 34.0 & 41.9 & 51.7 & 8.2 & 25.3 & 36.2 & 51.0 \\
        STS$\dagger$~\cite{wang2020statistic} & R(2+1)D & 38.1 & 58.9 & 68.9 & 77.2 & 16.4 & 36.9 & 50.5 & 65.4 \\
        CoCLR$\dagger$~\cite{han2020self} & S3D & 53.3 & 69.4 & 76.6 & 82.0 & 23.3 & 43.2 & 53.5 & 65.5 \\
        \hdashline
        \textbf{Ours} & R(2+1)D & 55.6 & 70.1 & 77.4 & 83.1 & 24.4 & 45.1 & 54.5 & 66.4 \\
        \hline
    \end{tabular}
    \caption{Results on video retrieval downstream task. We report R@k (k=1,5,10,20), $\dagger$ means pretrained with RGB and optical flow.}
    \label{tab:retrieval}
\end{table}

\noindent\textbf{Video Retrieval.}
We show the performance on video retrieval with R@k in Table~\ref{tab:retrieval}. All models are pretrained on UCF-101 with resolution $112\times 112$ for fair comparison. Generally, our method achieves superior results over both RGB-only and RGB-Flow two-stream methods, especially when k is small. It indicates that our method encodes desired characteristics into a more compact manifold.

\subsection{Concept Analysis}
Intuitively, actions can be represented by some general concepts, and the detailed feature description of these concepts help to discriminate similar action classes. To this end, in this section, we reveal how the learned static and dynamic concepts influence downstream action recognition.

\begin{table}
    \centering
    \begin{tabular}{c|ccccccc}
    \hline
        Feature & $\bm{v}$ & $\bm{q_v}$ & $\bm{q_v^s}$ & $\bm{q_v^d}$ & $\bm{F_v}$ & $\bm{F_v^s}$ & $\bm{F_v^d}$ \\
        \hline
        UCF-101 & $\text{ }$72.1$\text{ }$ & $\text{ }$66.3$\text{ }$ & $\text{ }$61.4$\text{ }$ & $\text{ }$62.6$\text{ }$ & $\text{ }$72.7$\text{ }$ & $\text{ }$68.3$\text{ }$ & $\text{ }$69.8$\text{ }$ \\
        HMDB-51 & 45.9 & 43.8 & 42.9 & 40.1 & 46.3 & 45.7 & 44.2 \\
        Diving-48 & 73.4 & 59.4 & 26.7 & 64.8 & 72.5 & 31.1 & 74.1 \\
        \hline
    \end{tabular}
    \caption{Results of static and dynamic concept analysis. The models in first two rows and the third row are respectively pretrained on Kinetics-400 and Diving-48.}
    \label{tab:concept}
\end{table}

\begin{figure}[t]
    \centering
    \includegraphics[width=0.9\linewidth]{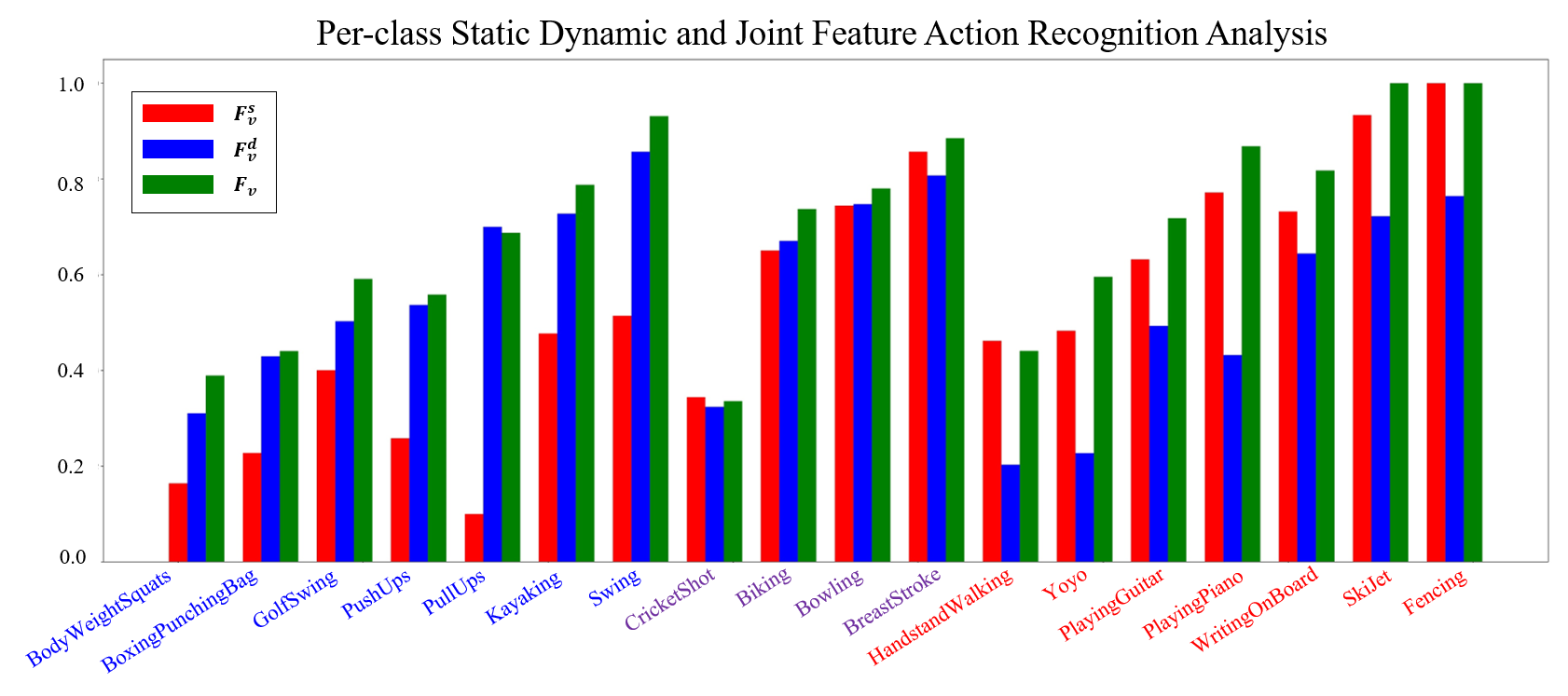}
    \caption{Per-class action recognition accuracy analysis. We compare the performance of using static, dynamic and joint concept related local feature set, namely $\bm{F_v^s}$, $\bm{F_v^d}$, $\bm{F_v}$.}
    \label{perclass}
\end{figure}

\begin{figure}
    \centering
    \subfigure[Static concept code]{
    \includegraphics[width=0.4\linewidth]{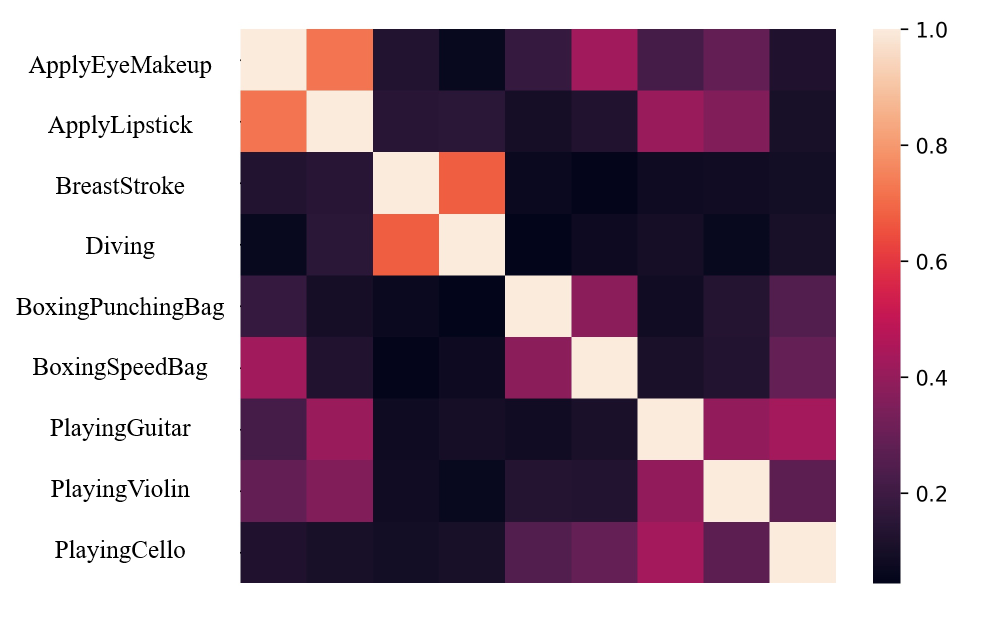}
    \label{heatmap-static}}
    \subfigure[Dynamic concept code]{
    \includegraphics[width=0.4\linewidth]{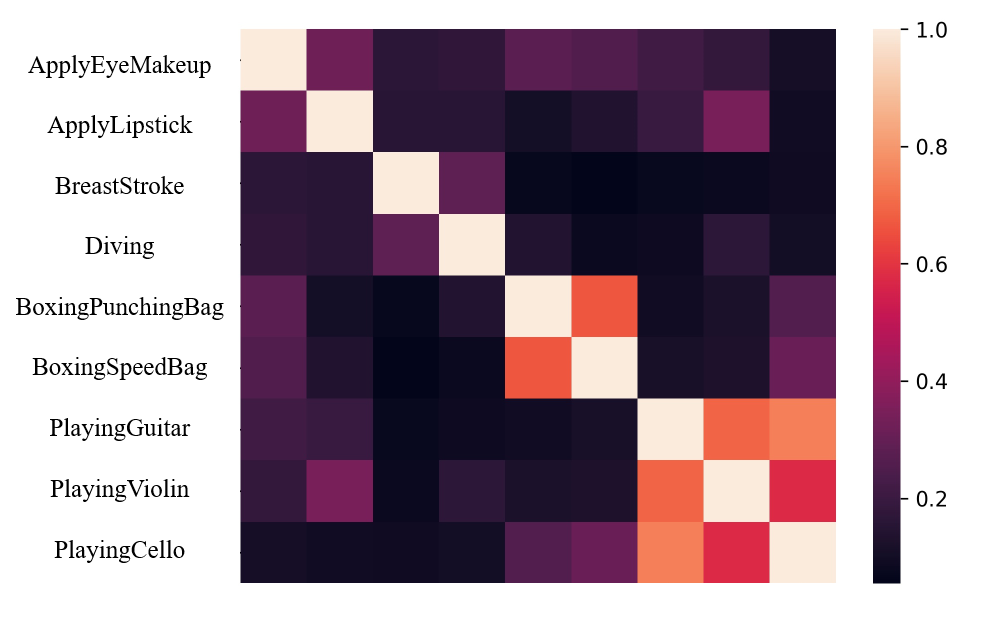}
    \label{heatmap-dynamic}}
    \caption{Decoupled concept code similarity. We respectively average static and dynamic concept latent codes, $\bm{q_v^s}$ and $\bm{q_v^d}$, within each category, then calculate cosine similarity.}
    \label{fig:heatmap}
\end{figure}

\noindent\textbf{Decoupled Concept for Action Recognition.} We first quantitatively analyze the static and dynamic concepts and their relevant local features on action recognition. We adopt different outputs from our learning framework and pass them through linear classifier to do action classification on UCF-101, HMDB-51 and Diving-48. Specifically, in default evaluation settings, we use the global average pooled $\bm{v}$ as input to the classifier. We also compare using the concept latent codes or the local feature set for recognition. Note that when using the local feature set, e.g., $\bm{F_v}$, we first filter out Top-10\% concepts from $\bm{q_v}$, then average the corresponding local features for classification. From Table~\ref{tab:concept}, we have several observations. First, using concept latent code for classification leads to performance drop, while the local feature set slightly improves performance. This is because we learn limited number of general concepts and could lose detailed information. While the local feature set effectively aggregates detailed information and drops redundant features, which helps to improve action recognition. Second, on UCF-101 and HMDB-51, static and dynamic concepts are almost of equal significance, and jointly utilizing static and dynamic concepts leads to best performance. Third, the dynamic concepts dominate action recognition on Diving-48, and the static concepts are nearly useless as expected. This is because different diving classes share the same background scene and only differ in motions, the static concepts could disturb motion pattern discrimination.

To further analyze the impact of the decoupled concepts on specific action categories, we select some typical classes from UCF-101 and visualize the per-class accuracy under different settings in Fig.~\ref{perclass}. Among the selected action categories, the blue ones are highly dominated by motions, the red ones may have ambiguous motion patterns but can be easily recognized by appearance, and for the purple ones, both static appearance and dynamic motion are discriminative. The per-class accuracy with different feature input is in line with our expectations, which indicates the decoupled concepts respectively reveal static and dynamic attributes. Besides, we analyze the inter-class similarity in static and dynamic concept latent space. In Fig.~\ref{fig:heatmap}, we visualize the similarity between different actions. Intuitively, some actions share similar background but with different motions, e.g., breaststroke and diving, while some possess similar movement but diverse appearances, e.g., playing different instruments. As expected, the former ones have higher inter-class similarity in static concept space while the latter ones are more similar regarding to dynamic concepts. 

\begin{figure}
    \centering
    \subfigure[Playing Violin]{
    \includegraphics[width=0.15\linewidth]{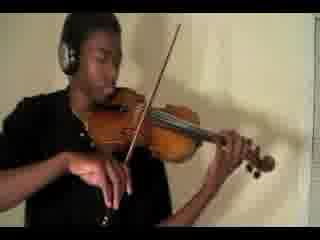}
    \includegraphics[width=0.15\linewidth]{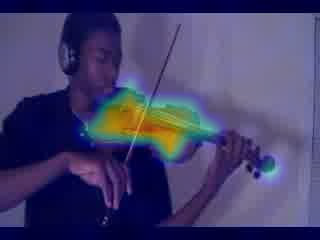}
    \includegraphics[width=0.15\linewidth]{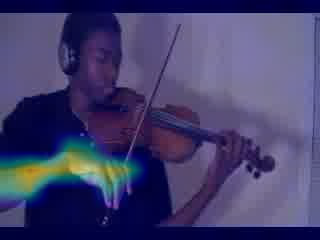}
    \label{violin}}
    \subfigure[Breast Stroke]{
    \includegraphics[width=0.15\linewidth]{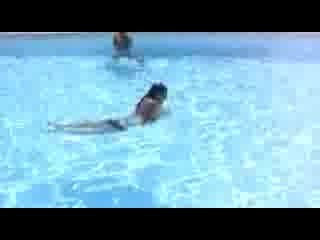}
    \includegraphics[width=0.15\linewidth]{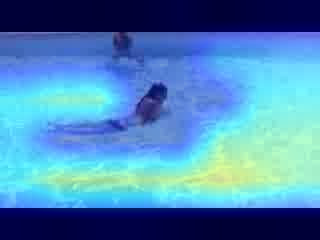}
    \includegraphics[width=0.15\linewidth]{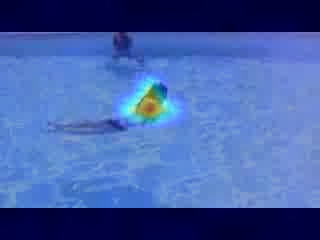}
    \label{swim}}
    \subfigure[Playing Cello]{
    \includegraphics[width=0.15\linewidth]{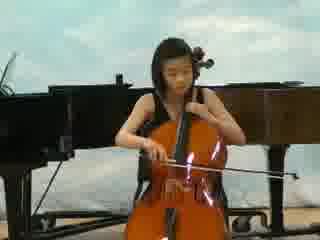}
    \includegraphics[width=0.15\linewidth]{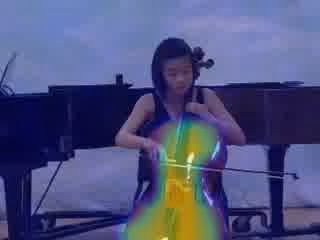}
    \includegraphics[width=0.15\linewidth]{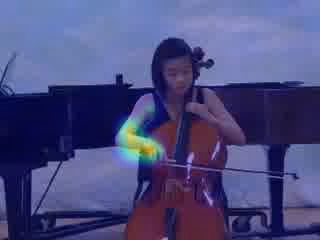}
    \label{cello}}
    \subfigure[Diving]{
    \includegraphics[width=0.15\linewidth]{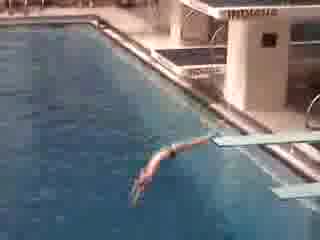}
    \includegraphics[width=0.15\linewidth]{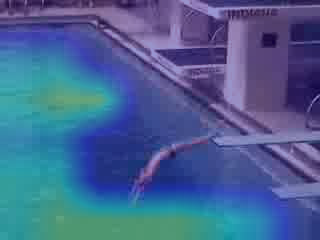}
    \includegraphics[width=0.15\linewidth]{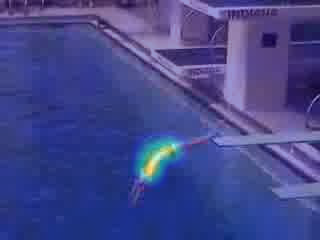}
    \label{dive}}
    \caption{Visualization of static and dynamic concept attention maps. Each subfigure left to right is: original frame, static concept attention map, dynamic concept attention map.}
    \label{attention2}
\end{figure}

\noindent\textbf{Visualization Results.}
For each clip, we respectively select a static and a dynamic concept with highest activation in latent space, and visualize the attention maps. Generally, the selected static concept attends to foreground objects or representative scene components, while the selected dynamic concepts highlights discriminative motions. Comparing Fig.~\ref{violin} and Fig.~\ref{cello}, they share similar dynamic attributes, i.e., almost synchronized forearm movements, but can be discriminated by static objects. Regarding to Fig.~\ref{swim} and Fig.~\ref{dive}, they happen in similar pools, but the dynamic concept helps to figure out distinct motion patterns. It reveals that the we learn meaningful static and dynamic concepts that focus on different aspects, these two jointly facilitate video understanding.

\subsection{Ablation Study}

We perform ablation studies on the loss function designs and crucial hyper-parameters. More ablative experiments please refer to Supplementary Material.

\begin{table}[]
    \centering
    \begin{tabular}{cccc|cc|cc}
    \hline
        \multirow{2}{*}{$\mathcal{L}_{aln}$} & \multirow{2}{*}{$\mathcal{L}_{fid}$} & \multirow{2}{*}{$\mathcal{L}_{div}$} & \multirow{2}{*}{$\mathcal{L}_{loc}$} & \multicolumn{2}{c|}{UCF-101} & \multicolumn{2}{c}{HMDB-51} \\
        \cline{5-8}
         & & & & $\text{ }$Linear$\text{ }$ & $\text{ }$Finetune$\text{ }$ & $\text{ }$Linear$\text{ }$ & $\text{ }$Finetune$\text{ }$ \\
        \hline
        \checkmark &  &  & & 61.4 & 76.3 & 40.3 & 44.7 \\
        \checkmark & \checkmark & \checkmark & & 68.1 & 80.1 & 43.2 & 47.9 \\
        \checkmark &  &  & \checkmark & 67.4 & 78.9 & 43.3 & 46.4 \\
        \checkmark & \checkmark & \checkmark & \checkmark & 72.1 & 82.1 & 45.9 & 49.7 \\
        \hline
    \end{tabular}
    \caption{Ablation study on loss functions. We pretrain on Kinetics-400.}
    \label{tab:loss}
\end{table}

\noindent\textbf{Overall Framework.} We first validate the effectiveness of the loss functions in Table~\ref{tab:loss}. The model is pretrained with default concept numbers, $K_s=K_d=50$, and the decoupled concept alignment objective $\mathcal{L}_{aln}$ serves as the baseline. We can observe that the two regularizations $\mathcal{L}_{fid}$ and $\mathcal{L}_{div}$ significantly improve the performance. This is because these two terms reduce redundancy in the learned concept prototypes and contribute to more compact and diverse concept formulation, which effectively guides representation learning. And regarding to $\mathcal{L}_{loc}$, it also brings significant improvement since this objective explicitly contrasts local features of valid concepts and facilitates detailed local feature perception.

\begin{table}[]
    \centering
    \begin{tabular}{cc|cc|cc}
    \hline
        \multirow{2}{*}{$\text{ } K_s\text{ }$} & \multirow{2}{*}{$\text{ } K_d\text{ }$} & \multicolumn{2}{c|}{UCF-101} & \multicolumn{2}{c}{HMDB-51} \\
        \cline{3-6}
         & & $\text{ }$w/ $\mathcal{L}_{loc}\text{ }$ & $\text{ }$w/o $\mathcal{L}_{loc}$$\text{ }$ & $\text{ }$w/ $\mathcal{L}_{loc}\text{ }$ & $\text{ }$w/o $\mathcal{L}_{loc}$$\text{ }$ \\
        \hline
        25 & 25 & 70.3 & 61.2 & 43.0 & 39.4 \\
        25 & 50 & 71.7 & 66.3 & 44.1 & 40.8 \\
        50 & 25 & 71.3 & 65.2 & 44.8 & 42.4 \\
        50 & 50 & 72.1 & 68.1 & 45.9 & 43.2 \\
        100 & 100 & 72.3 & 68.8 & 45.8 & 44.3 \\
        200 & 200 & 72.3 & 69.4 & 45.6 & 44.1 \\
        \hline
    \end{tabular}
    \caption{Ablation study on concept numbers. We report linear probe accuracy.}
    \label{tab:number}
\end{table}

\noindent\textbf{Number of Concepts.} We also explore the impact of different concept numbers in Table~\ref{tab:number}. With the help of $\mathcal{L}_{loc}$, the performance slightly improves when $K_s$ and $K_d$ increases, and maintains stable in range of 50 to 200. While without $\mathcal{L}_{loc}$, the performance dramatically drops when the concept numbers become small. Because when $K_s$ and $K_d$ are small, the latent space captures general concepts but loses detailed information to discriminate similar actions. But when combined with $\mathcal{L}_{loc}$, the model adaptively attends to detailed local features with desired concepts, which makes up for the information loss to a large extent.

\section{Conclusion}

In this paper, we propose to learn general static and dynamic visual concepts to guide self-supervised video representation learning. We design decoupled concept alignment objective with regularizations to jointly optimize feature representations and concept distributions. Then we refer to the learned concepts to aggregate detailed local features corresponding to different concepts. We utilize the concept latent code to filter out redundant concepts with low activations, and perform concept-level local feature contrast for detailed video understanding. We achieve state-of-the-art results on UCF-101, HMDB-51 and Diving-48. The ablation studies demonstrate that the integration of general concept learning and detailed local feature contrast improves video representation learning.

\noindent\textbf{Acknowledgement.}
This work is supported by GRF 14205719, TRS T41-603/20-R, Centre for Perceptual and Interactive Intelligence, and CUHK Interdisciplinary AI Research Institute.

\clearpage
%
%
\bibliographystyle{splncs04}
\bibliography{egbib}
\end{document}